\documentclass[letterpaper, 10 pt, conference]{ieeeconf}  

\IEEEoverridecommandlockouts                              

\usepackage{graphicx} 
\usepackage{epstopdf} 
\usepackage{mathptmx} 
\usepackage{times} 
\usepackage{amsmath,amssymb,amsfonts}
\usepackage{multirow}
\usepackage[dvipsnames]{xcolor}
\usepackage{caption}
\usepackage{subcaption}
\usepackage{booktabs}
\usepackage{subfig}
\usepackage{array}

\newcolumntype{P}[1]{>{\centering\arraybackslash}p{#1}}

\title{\LARGE \bf
Pushing Radar Odometry Beyond the Pavement: Current Capabilities and Challenges
}

\author{Shaunak Kolhe$^{1}$, Peng Jiang$^{1}$, Maggie Wigness$^2$, Philip Osteen$^2$,  \\ 
Timothy Overbye$^2$, Christian Ellis$^2$ and Srikanth Saripalli$^{1}$
\thanks{ This paper was created by 1. Texas A\&M University and 2. DEVCOM Army Research Laboratory. Texas A\&M University is at College Station, TX, USA. DEVCOM Army Research Laboratory is at Adelphi, MD, USA %
        {\tt\small kolheshaunak@tamu.edu}, %
        {\tt\small maskjp@tamu.edu}, %
        {\tt\small ssaripalli@tamu.edu}}}%

\begin{document}

\maketitle
\thispagestyle{empty}
\pagestyle{empty}

\begin{abstract}


Radar offers unique advantages for localization in unstructured environments, including robustness to weather, lighting, and airborne particulates. While most prior work has studied radar odometry in urban, largely planar settings, its performance in off-road environments remains less understood. In this paper, we investigate the potential of radar for off-road odometry estimation and identify key challenges that arise from full $SE(3)$ vehicle motion, terrain-induced ground returns, and sparse or unstable features. To address these issues, we introduce two simple baselines: Radar-KISSICP, which applies motion compensation to generate 3D-aware radar pointclouds, and Radar-IMU, which leverages IMU preintegration to stabilize scan matching. Experiments on the Great Outdoors (GO) dataset demonstrate that these baselines improve trajectory estimation in challenging routes and provide a reference point for future development of radar odometry in off-road robotics.

\end{abstract}

\section{INTRODUCTION}

Robust localization is a fundamental capability for autonomous robots operating in diverse and unstructured environments. While cameras and LiDARs have long been the dominant sensing modalities for odometry, their performance degrades severely under adverse weather or visually degraded conditions such as rain, snow, fog, dust, or smoke. Millimeter-wave radar, with its longer wavelengths and Doppler measurement capability, has recently re-emerged as a compelling alternative \cite{harlow_new_2024}. Unlike light-based sensors, radar can penetrate particulates and provide reliable returns at long range, enabling continued operation when other modalities degrade or fail.

Encouraged by these properties, a growing body of work has explored radar odometry and mapping in automotive and urban driving scenarios. Recent methods, such as CFEAR \cite{adolfsson_lidar_level_2023} and ORORA \cite{lim_orora_2023}, achieve near LiDAR-level localization performance on benchmark datasets like Oxford RobotCar~\cite{barnes_oxford_2020}, MulRan~\cite{kim_mulran_2020} and Boreas~\cite{burnett_boreas_2023}. However, these datasets primarily constrain vehicle motion to planar trajectories in $SE(2)$. As a result, many state-of-the-art radar odometry systems implicitly rely on the assumption of flat terrain and limited pitch or roll excitation, which does not generalize to off-road conditions.

In contrast, off-road environments demand full $SO(3)$ motion estimation due to rough terrain, elevation changes, and frequent loss of stable landmarks. These conditions expose radar-specific failure modes: ground returns become confounding rather than outliers, false correspondences accumulate in ICP pipelines, and sparse radar pointclouds lead to ill-conditioned registration. To address these challenges, we evaluate existing radar odometry engines in off-road settings and propose two extensions: (i) Radar-KISSICP, which is based on KISS-ICP\cite{vizzo_ral23} and incorporates a lightweight $SO(3)$ motion compensation step to produce sparse but 3D-aware radar pointclouds, and (ii) Radar-IMU, which leverages IMU preintegration to provide a robust initial guess for radar scan-matching in $SE(3)$.

Through experiments on the Great Outdoors (GO) dataset~\cite{jiang2025gogreatoutdoorsmultimodal}, we show that our localization methods can match or outperform existing approaches in challenging off-road environments, including scenarios with severe pitch/roll and terrain discontinuities. We further analyze failure cases, such as ravine traversals, where radar-only pipelines accumulate significant drift, and demonstrate how IMU integration helps recover accurate trajectories. Our evaluation and results emphasize both the promise and the limitations of radar odometry in unstructured terrain, laying the groundwork for future robust off-road localization pipelines.

\begin{figure}
    \centering
    \includegraphics[width=\linewidth]{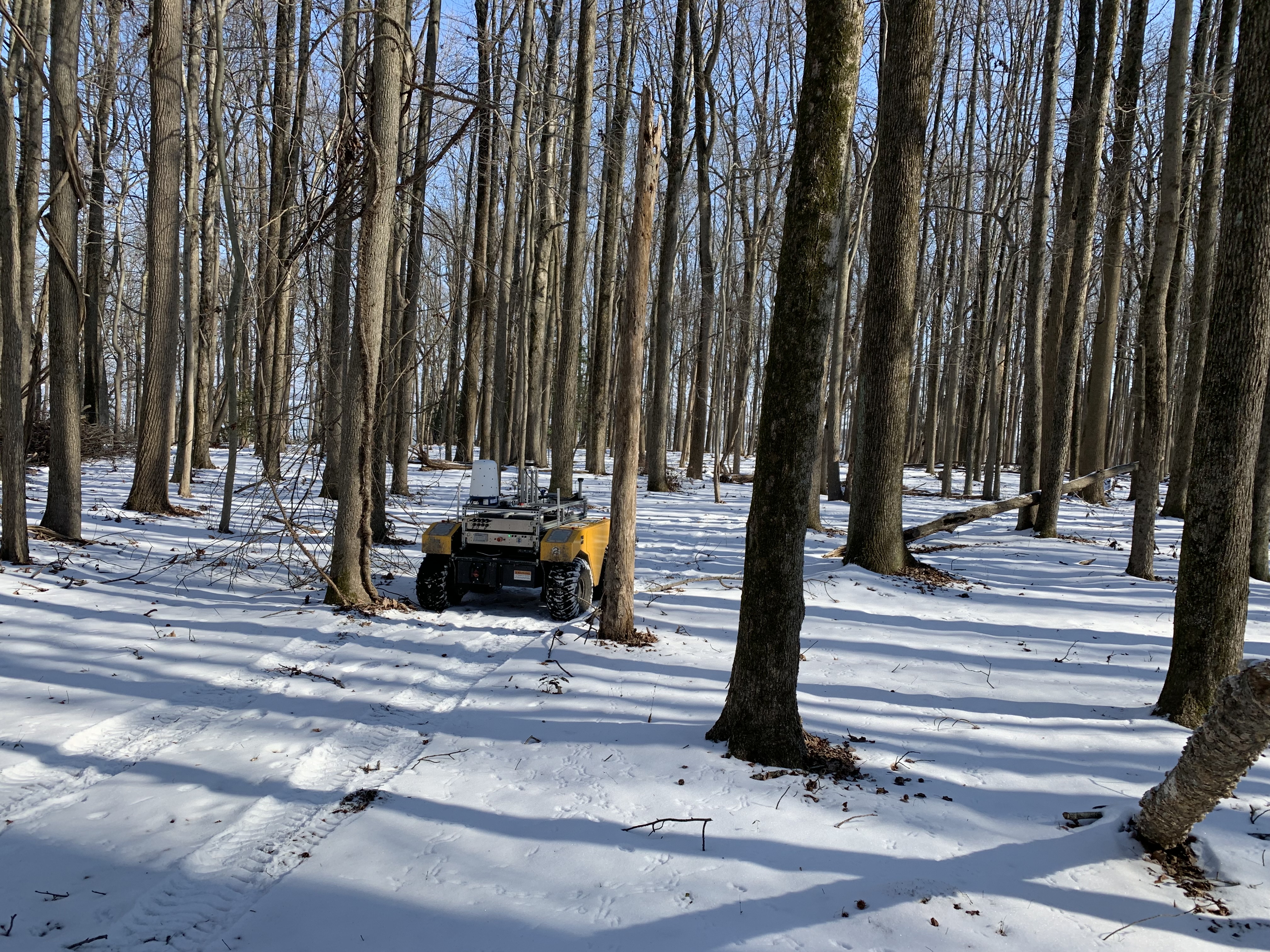}
    \caption{\small Motivating scenario for the use of radar in an off-road, degraded environment. A Clearpath Warthog equipped with a Navtech radar operates in a snow covered dense forest.}
    \label{fig:placeholder}
    \vspace{-20pt}
\end{figure}
\section{Related Work}
Early radar odometry approaches focused on adapting classical registration techniques to handle measurement noise and sparsity. 
Kung et al. \cite{kung_normal_2021} proposed a normal distributions transform (NDT) based RADAR odometry that also accounts for uncertainties through a probabilistic submap, outperforming state-of-the-art automotive and learning-based scanning methods by better handling outliers. Rapp et al. \cite{rapp_probabilistic_2017} also applied joint doppler-clustering based NDT, weighting each RADAR point by similarity between measured and estimated velocities to reduce outlier effects. 
To address the sparsity issue in automotive RADAR, Holder et al. \cite{holder_iv19} proposed submap matching, combining several RADAR scans to construct more information, though this method did not consider the uncertainty of ego-motion or RADAR measurements.

More recently, Adolfsson et al. presented the CFEAR approach~\cite{adolfsson_lidar_level_2023}, which uses a feature extraction algorithm to generate filtered surface points that can be reliably scan-matched from frame to frame, demonstrating LIDAR-level localization with RADAR. Lim et al. developed ORORA \cite{lim_orora_2023}, an outlier-robust RADAR odometry method that demonstrated robust performance even with numerous outliers and was less sensitive to the specific feature extraction method used, particularly in featureless scenes. 

Direct methods, on the other hand, directly compare or use raw RADAR measurements without explicit feature extraction.
Kellner et al. \cite{kellner} used the angular position and radial velocity of automotive RADAR sensors to estimate ego-motion, classifying stationary targets with RANSAC to fit velocity estimates.
Le Gentil et al. \cite{gentil_dro_nodate} introduced a Direct RADAR Odometry (DRO) method that is notable for being the first direct RADAR registration method to account for continuous motion and Doppler-based distortion in a gradient-based optimization, performing well in both structured and unstructured environments, including featureless tunnels.

Recent advancements have leveraged deep learning to extract robust features and improve RADAR odometry performance, often outperforming hand-crafted methods.
Barnes et al. \cite{barnes_under} demonstrated a fully differentiable, correlation-based RADAR odometry pipeline that learns a binary mask to remove distractor features, such as objects in motion and RADAR noise.
Other approaches use unsupervised learning, such as the HERO method by Burnett et al. \cite{burnett_rss21}, which combines probabilistic trajectory estimation and deep learned features without requiring ground truth pose information for training.
This unsupervised method approaches the method in \cite{adolfsson_lidar_level_2023}, outperforms the supervised method in \cite{barnes_under} and demonstrates effectiveness under adverse weather conditions.

Lisus et al. \cite{lisus2025pointingwayrefiningradarlidar} further advanced RADAR-LIDAR localization by presenting a deep-learning-based approach that learns to weight RADAR points for ICP, effectively filtering out harmful RADAR artifacts, noise, and even other vehicles, thus improving convergence and reducing localization errors.
Lisus et al. \cite{lisus_are_2025} investigated whether Doppler velocity measurements are useful for spinning RADAR odometry and presented a novel method to analytically extract these velocities while retaining full angular resolution. Their experiments demonstrated that using Doppler measurements improves performance in easy geometric environments and maintains good performance in geometrically degenerate situations like tunnels or skyways where other methods might fail. 

While the above demonstrate the effectiveness of RADAR as a robust odometry source, rarely are these methods tested in offroad environments. It is hence in our interest to verify if this performance is applicable in more aggressive terrain, and identify the challenges therein.
\begin{figure}[t]
    \centering
    \setlength{\tabcolsep}{0pt}
    \begin{tabular}{P{0.5\columnwidth}  P{0.5\columnwidth}}
        \includegraphics[height=2in]{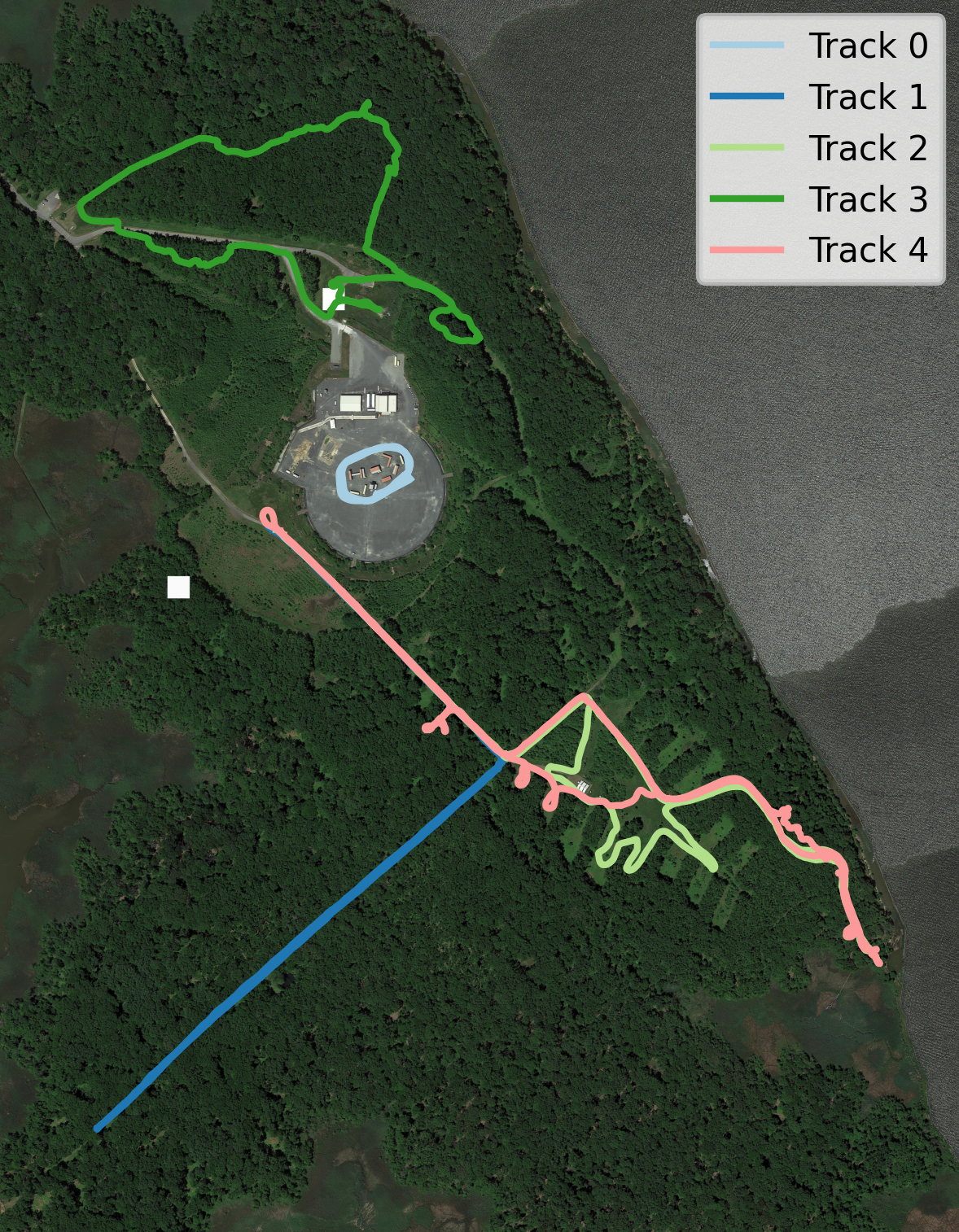} &
        \includegraphics[height=2in]{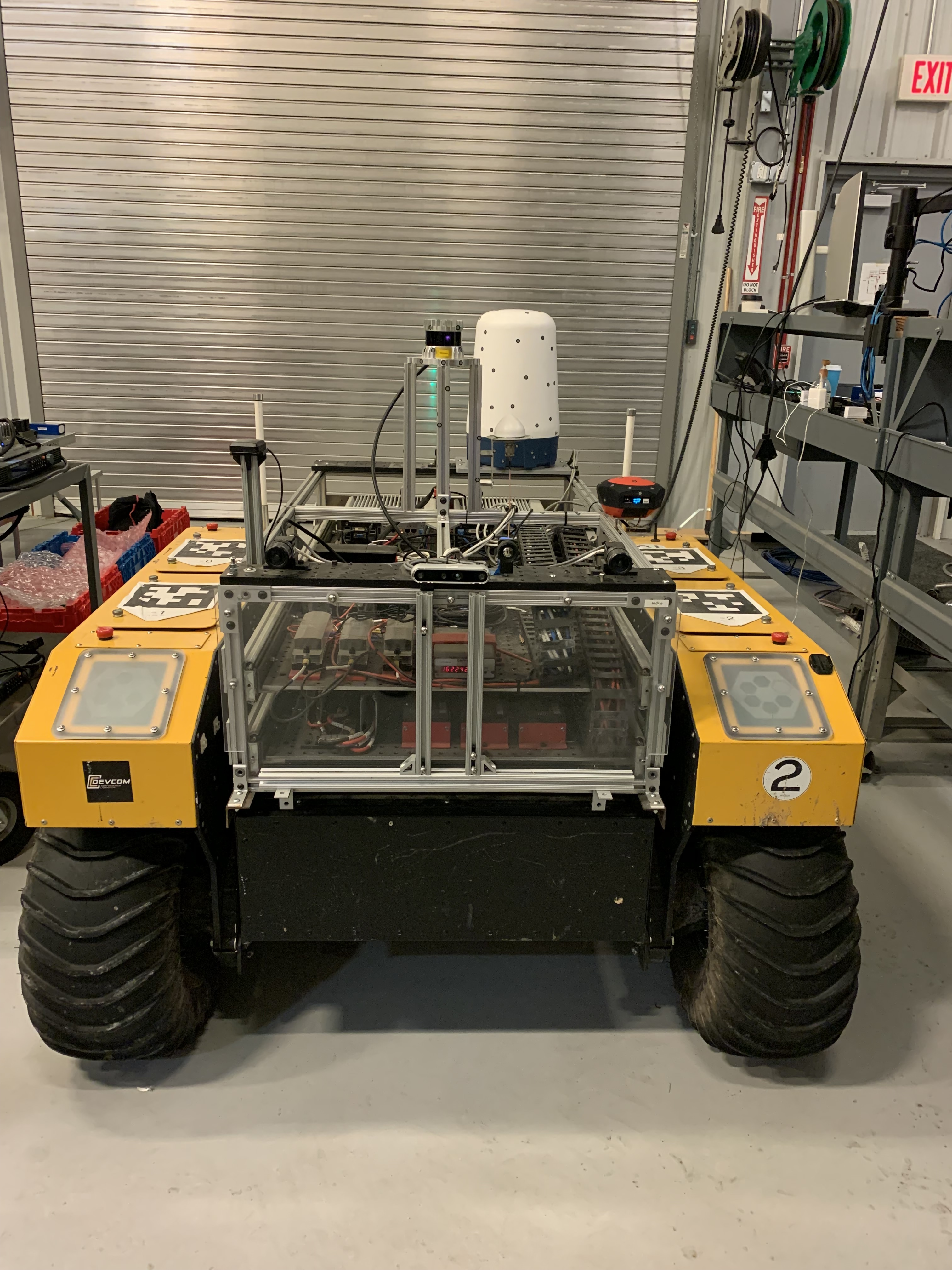} 
    \end{tabular}
    \caption{Overhead view of routes from the GO dataset~\cite{jiang2025gogreatoutdoorsmultimodal} (left) and the Warthog sensor setup used for the dataset (right).}
    \label{fig:GO_map}
\vspace{-10pt}
    
\end{figure}

\section{Benchmarking Framework}

To systematically study radar odometry in off-road environments, we design a benchmarking framework that evaluates state-of-the-art radar odometry engines as well as two diagnostic baselines. Our goal is to reveal the strengths and limitations of existing methods when vehicle motion extends beyond planar $SE(2)$ into full $SE(3)$.
\vspace{-3pt}
\subsection{Odometry Engines Under Evaluation}

We benchmark two representative feature-based radar odometry methods that have demonstrated competitive performance in automotive datasets: CFEAR~\cite{adolfsson_lidar_level_2023} and ORORA~\cite{lim_orora_2023}. In addition, we introduce two lightweight extensions, Radar-KISSICP and Radar-IMU, designed as \textit{diagnostic baselines} to probe failure cases in off-road conditions. A brief description of each method is provided below.

\subsubsection{CFEAR} 
The CFEAR pipeline uses a surfel-based feature extraction method and a robust point-to-line error optimization approach for pose estimation. Each radar sweep is filtered to retain the $K$-strongest intensities per azimuth and projected into Cartesian space. Motion compensation is then applied using the most recent velocity estimates by reprojecting each point into the time $t$ of the sweep center. Oriented surface points are extracted via eigen-decomposition to compute local normals, and registration is performed against a keyframe using a point-to-line error metric with a Cauchy loss using ~\cite{noauthor_ceres_nodate}. Following~\cite{adolfsson_lidar_level_2023}, we adopt a configuration of 50 keyframes with the Cauchy loss, which was reported to yield the best performance.

\subsubsection{ORORA} 
The ORORA pipeline addresses outlier sensitivity in feature-based radar odometry by introducing a decoupling-based matching strategy. Features are extracted using the Cen2018 detector~\cite{cen_precise_2018}, described with ORB~\cite{rublee_orb_2011}, and matched via brute-force search with a ratio test~\cite{lowe_object_1999}. Doppler compensation is applied to the range measurements, and a maximum clique inlier selection~\cite{rossi_parallel_2013} is used to prune spurious correspondences. The final relative transform is estimated by solving for rotation using a graduated non-convexity method, followed by anisotropic component-wise estimation of translation.

\subsubsection{Radar-KISSICP} 
To analyze failures of CFEAR and ORORA in off-road settings, we adapt KISS-ICP~\cite{vizzo_ral23} to radar data. Radar features detected by Cen2018 are cast into a pointcloud, followed by a rotation-only motion compensation step that produces sparse but 3D-aware radar pointclouds. Registration is then performed using point-to-point ICP. While this reduces the planar assumption, the resulting sparsity makes the ICP problem ill-conditioned: incorrect nearest-neighbor associations often dominate, and convergence is highly sensitive to the initial guess.

\subsubsection{Radar-IMU} 
To mitigate the sensitivities of Radar-KISSICP, we extend the approach by incorporating IMU preintegration~\cite{imu_preintegration} using GTSAM~\cite{gtsam} to provide an initial $SE(3)$ pose estimate for ICP. This inertial prior introduces vertical displacement awareness otherwise unobservable from radar-only motion models, reducing false map features and improving recovery from drift. Importantly, odometry still updates at radar frame rate, and no bias correction is applied. We refer to this variant as Radar-IMU throughout the paper.
\begin{table}
    \tiny
    \centering
    \begin{tabular}{c|ccccc}
    \hline
         Dataset & Manufacturer & Model & Range Resolution & Range \\
    \hline
        MulRan \cite{kim_mulran_2020} & Navtech & CIR204-H  & 0.06 m & 200 m \\
        Boreas \cite{burnett_boreas_2023} & Navtech & CIR304-H  & 0.04 m & 250 m \\
        Oxford Radar Dataset \cite{barnes_oxford_2020} & Navtech & CTS350-X & 0.04 m & 163 m \\
        GO Dataset \cite{jiang2025gogreatoutdoorsmultimodal} & Navtech & CTS350-X & 0.04 m & 270 m \\
    \hline
    \end{tabular}
    \caption{\small Radar sensor specifications from the datasets under test (all radar have $360^{\circ}$ FOV, $0.9^{\circ}$ azimuth resolution)}
    \vspace{-10pt}
    \label{tab:rad_types}
\end{table}

\subsection{Datasets}
We evaluate across both automotive and off-road datasets to expose the differences between planar $SE(2)$ motion and full $SE(3)$ motion:

\begin{itemize}
    \item \textbf{Automotive datasets:} Oxford Radar RobotCar~\cite{barnes_oxford_2020}, Boreas~\cite{burnett_boreas_2023} and MulRan~\cite{kim_mulran_2020} were collected in structured, largely planar urban environments. These datasets serve as baseline environments where current radar odometry has traditionally achieved strong performance. We use the '2019-01-18-15-20-12' sequence from Oxford, the 'KAIST03' sequence from MulRan and the '2021-06-17-17-52' sequence from Boreas in our experiments. 
    
    \item \textbf{Off-road dataset:} The Great Outdoors (GO) dataset~\cite{jiang2025gogreatoutdoorsmultimodal}, was collected using a Clearpath Warthog UGV~\cite{noauthor_warthog_nodate} (Fig.~\ref{fig:GO_map}) along forest trails, ravines, and snow-covered terrain with frequent pitch/roll excitation. We focus on three representative routes with increasing difficulty:
    \begin{itemize}
        \item \textbf{Route~1:} A paved area with few distinctive features. The robot covered 0.58~km in 3.43~minutes, reaching a peak speed of 4.0~m/s.
        \item \textbf{Route~3:} A challenging mix of gravel trails and forested regions, where difficult terrain limited progress to an average speed of 0.98~m/s.
        \item \textbf{Route~4:} A trail run with multiple deviations into off-road areas before returning to the main path, resulting in an average speed of 1.59~m/s.
    \end{itemize}
\end{itemize}

\begin{figure}[t]
    \centering
    \setlength{\tabcolsep}{0pt}
    \begin{tabular}{c}
        \includegraphics[width=0.9\columnwidth]{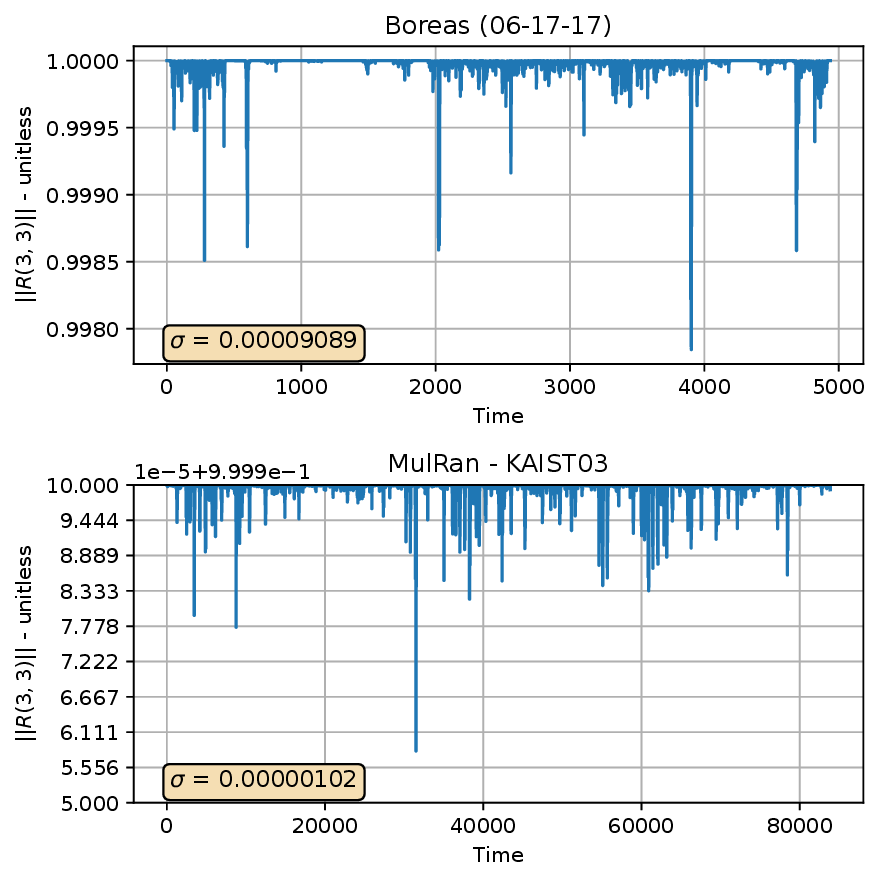} \\
        \small{(a) Overlap on $SE(2)$: Automotive Sequences} \\
        \includegraphics[width=0.9\columnwidth]{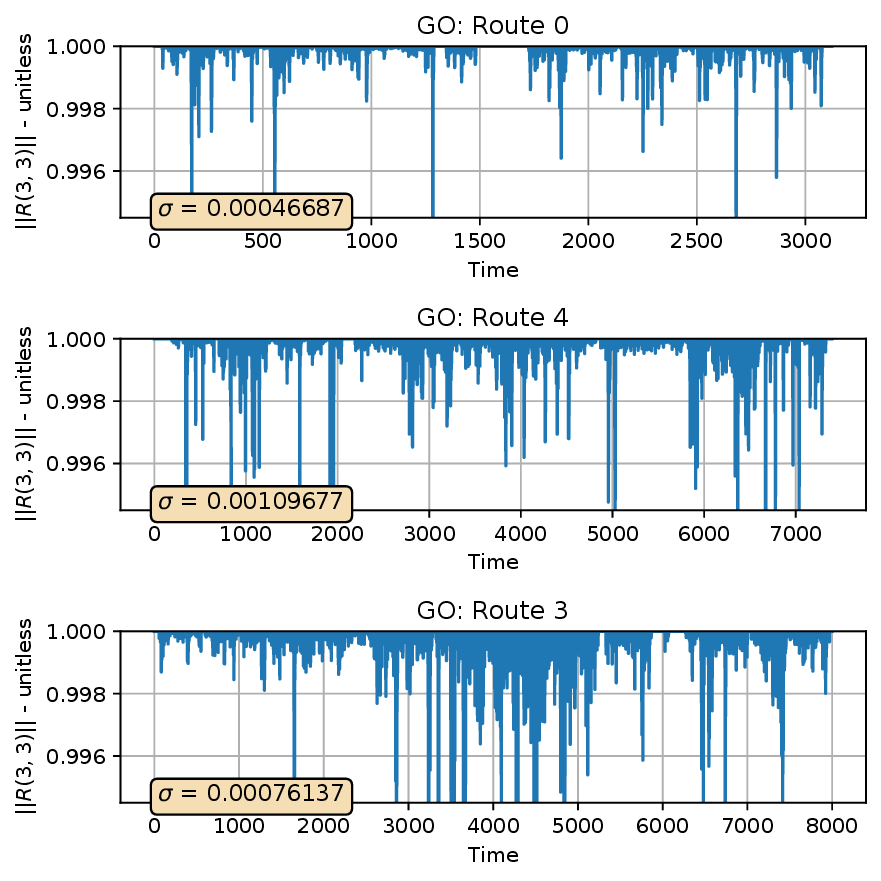}\\
        \small{(b) Overlap on $SE(2)$ Offroad Sequences}
    \end{tabular}
    \caption{Overlap of relative transforms in $SE(3)$ on $SE(2)$. \\
    \small
     While neither automotive nor offroad sequences exhibit motion restricted to a plane, the deviation from this assumption is greater in the offroad sequences as can be seen from the standard deviation of the data from unity. }
    \label{fig:planarity}
    \vspace{-20pt}
\end{figure}

Table~\ref{tab:rad_types} summarizes the radar sensor specifications across datasets. To highlight the difference in motion regimes, we compute the overlap of ground-truth relative transforms with the XY-plane (Fig.~\ref{fig:planarity}). While automotive datasets exhibit nearly pure yaw motion, the GO sequences show frequent pitch and roll, underscoring the need for full $SE(3)$ evaluation.

\subsection{Evaluation Metrics and Setup}
We adopt standard odometry metrics following KITTI~\cite{geiger_vision_2013} and EVO~\cite{grupp2017evo}:
\begin{itemize}
    \item \textbf{Relative Pose Error (RPE):} Captures local drift over fixed-length segments.
    \item \textbf{Absolute Trajectory Error (ATE):} Measures accumulated global trajectory drift.
\end{itemize}

All methods are evaluated in full $SE(3)$, rather than restricting to $SE(2)$, to properly account for terrain-induced vertical motion. Experiments are conducted on an AMD Ryzen 7 7700X CPU at 5.3~GHz with 64~GB RAM, using CPU-only execution to ensure consistent runtime comparisons.

\begin{table}[]
    \centering
    \tiny
    \begin{tabular}{c||c|c|c|c|c}
    \toprule
      \textbf{Sequence} & \textbf{Method} & \textbf{Avg. trans.} & \textbf{Avg. rot.} & \textbf{ATE tra.} & \textbf{ATE rot.} \\
         \midrule
         \multirow{2}{*}{Oxford} & CFEAR & \textbf{1.50} & \textbf{0.01} & \textbf{15.48} & \textbf{0.12} \\
                            & ORORA & 3.51 & 0.04 & 45.06 & 0.14 \\
        \midrule
         \multirow{2}{*}{MulRan} & CFEAR & 87.77 & 1.33 & 72.61 & \textbf{1.28} \\
                            & ORORA & \textbf{52.20} & \textbf{1.06} & \textbf{36.23} & 1.29 \\
    \midrule
         \multirow{2}{*}{Boreas} & CFEAR & 1.79 & 0.03 & 81.84 & 0.15 \\
                            & ORORA & - & - & - & - \\
        \bottomrule
    \end{tabular}
    \caption{Quantitative results for current algorithms on automotive datasets. \\
    \small
    Relative translational and rotational error are reported using the KITTI~\cite{geiger_vision_2013} metrics. Also shown are absolute trajectory error for translation in \% and for rotation in rad/100m. Best value per metric per dataset in bold. Lower error values indicate better performance. We exclude results for ORORA on the Boreas dataset as it is not compatible to run.}
    \label{tab:automotive_metrics}
\end{table}

\begin{table}
    \tiny
    \centering
    \begin{tabular}{c||c|c|c|c|c}
    \toprule
         \textbf{Sequence} & \textbf{Method} & \textbf{Avg. trans.} & \textbf{Avg. rot.} & \textbf{ATE tra.} & \textbf{ATE rot.} \\
         \midrule
         \multirow{4}{*}{Route 1} & CFEAR         & 9.12 & 0.80 & 180.44 & 1.67 \\
                           & ORORA          & 5.95 & 0.52 & 127.59 & 1.43 \\
                           & Radar-KISSICP  & 1.47 & 0.44 & 41.17  & 1.35 \\
                           & Radar-IMU      & 1.06 & 0.46 & 21.69  & 2.00 \\
        \midrule
         \multirow{4}{*}{Route 3} & CFEAR         & 3.24 & 1.76 & 61.08 & 1.93 \\
                           & ORORA          & 4.19 & 1.64 & 25.88 & 1.95 \\
                           & Radar-KISSICP  & 3.59 & 1.70 & 50.97 & 1.97 \\
                           & Radar-IMU      & 2.89 & 1.57 & 15.12 & 1.56 \\
        \midrule
         \multirow{4}{*}{Route 4} & CFEAR   & 3.10  & 1.27 & 232.27 & 1.66 \\
                           & ORORA          & 46.66 & 1.65 & -- & -- \\
                           & Radar-KISSICP  & 1.67  & 1.53 & 42.32  & 2.04 \\
                           & Radar-IMU      & 12.48 & 1.46 & 54.45  & 1.60 \\
        \bottomrule

    \end{tabular}
    \caption{Quantitative results for candidate methods on Offroad Datasets. Relative translational and rotational error are reported using the KITTI~\cite{geiger_vision_2013} metrics. Also shown are absolute trajectory error for translation in \% and for rotation in rad/100m. Lower error values indicate better performance.}
    \label{tab:offroad_metrics}
    \vspace{-20pt}
\end{table}
\section{Quantitative and Qualitative Results}
\subsection{Automotive Datasets}
Table~\ref{tab:automotive_metrics} summarizes results on Oxford and MulRan. Both CFEAR and ORORA achieve competitive performance, consistent with prior reports. CFEAR generally produces lower translational error, while ORORA shows robustness to spurious correspondences. These experiments confirm that both methods are well-suited to planar $SE(2)$ motion, providing a baseline for comparison with off-road scenarios.

\subsection{Off-road Datasets}

Table~\ref{tab:offroad_metrics} presents results for Routes~1, 3, and 4 from the GO dataset. In contrast to automotive datasets, both CFEAR and ORORA show substantial drift under $SE(3)$ excitation. On Route~1, where terrain is relatively flat, performance is comparable to the current automotive approaches. However, on Routes~3 and 4, which involve ravines and trail deviations, performance degrades sharply. ORORA in particular fails to complete Route~4, while CFEAR accumulates large trajectory errors despite producing plausible local estimates.

\begin{figure}[h!]
    \centering
    \setlength{\tabcolsep}{0pt}
    \begin{tabular}{c}
        \includegraphics[width=0.9\columnwidth]{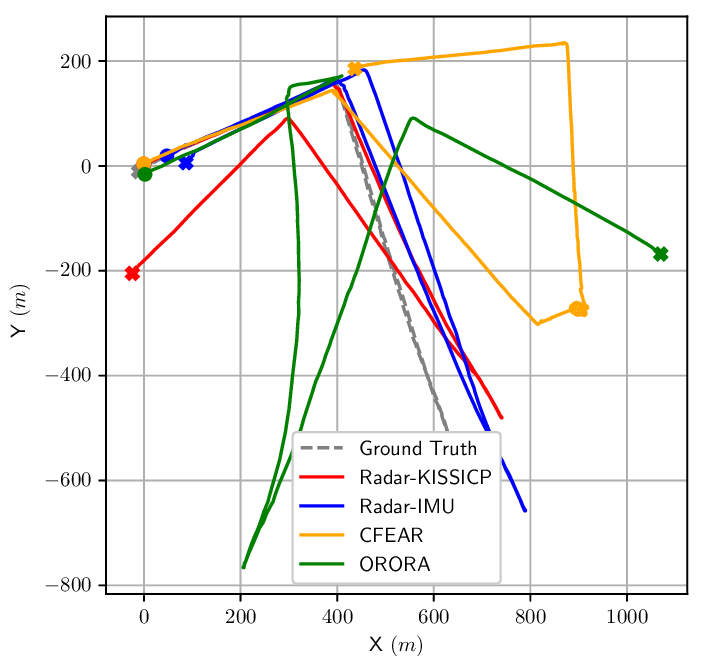}\\
        (a) Route 1 \\
        \includegraphics[width=0.9\columnwidth]{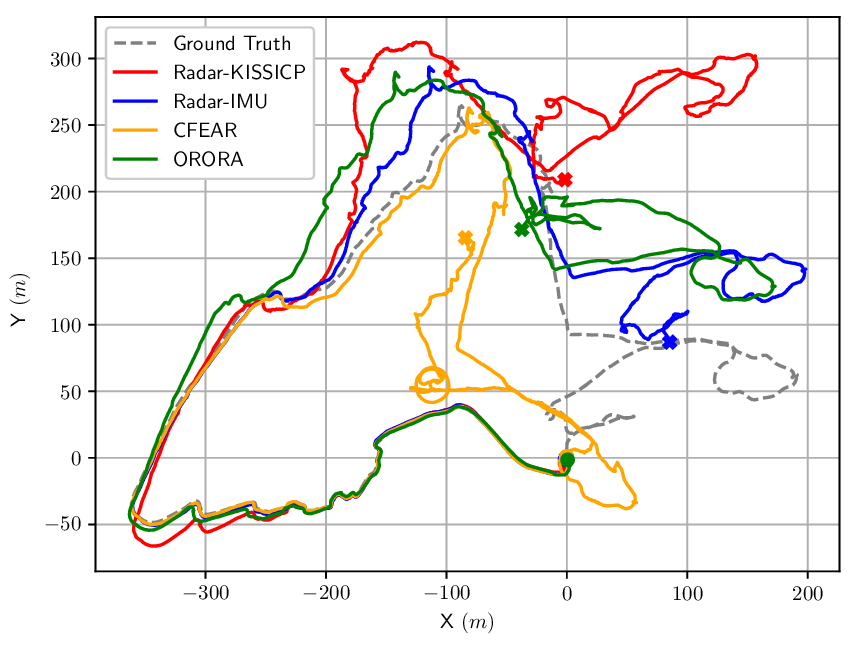} \\
        (b) Route 3 \\
        \includegraphics[width=0.9\columnwidth]{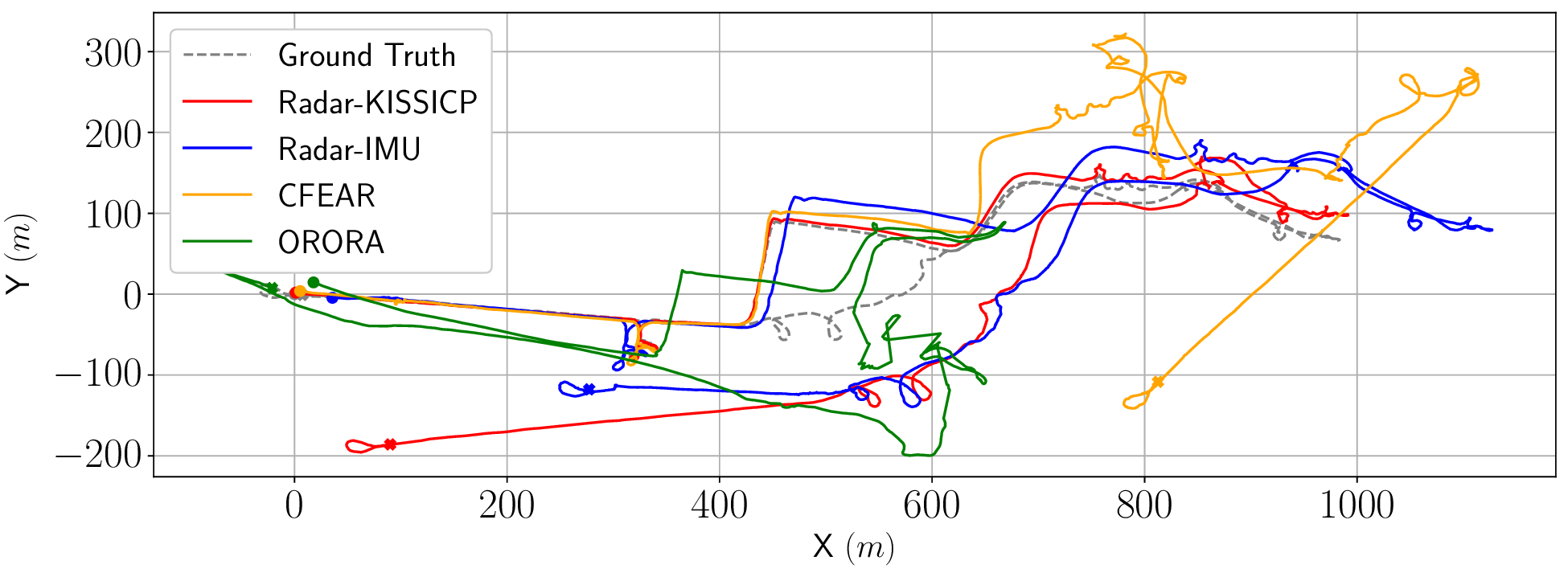}\\
        (c) Route 4 \\
    \end{tabular}
    \caption{Estimated trajectories from current RADAR odometry engines and our alternative methods}
    \label{fig:traj_offroad}
    \vspace{-20pt}
\end{figure}

Incorporating motion compensation (Radar-KISSICP) reduces absolute trajectory error in several cases, but the sparse nature of radar pointclouds leads to unstable nearest-neighbor associations. The addition of IMU preintegration (Radar-IMU) further improves robustness, particularly in the ravine section of Route~3, where it successfully recovers vertical displacement that radar-only models cannot observe (Fig.~\ref{fig:traj_offroad}). These improvements demonstrate the importance of inertial priors for $SE(3)$ motion, even in radar-dominant pipelines.

\begin{figure*}[!t]
    \centering
    \setlength{\tabcolsep}{0.2pt}
    \begin{tabular}{P{0.4\textwidth} P{0.6\textwidth}}
        \includegraphics[width=0.4\textwidth,height=2in]{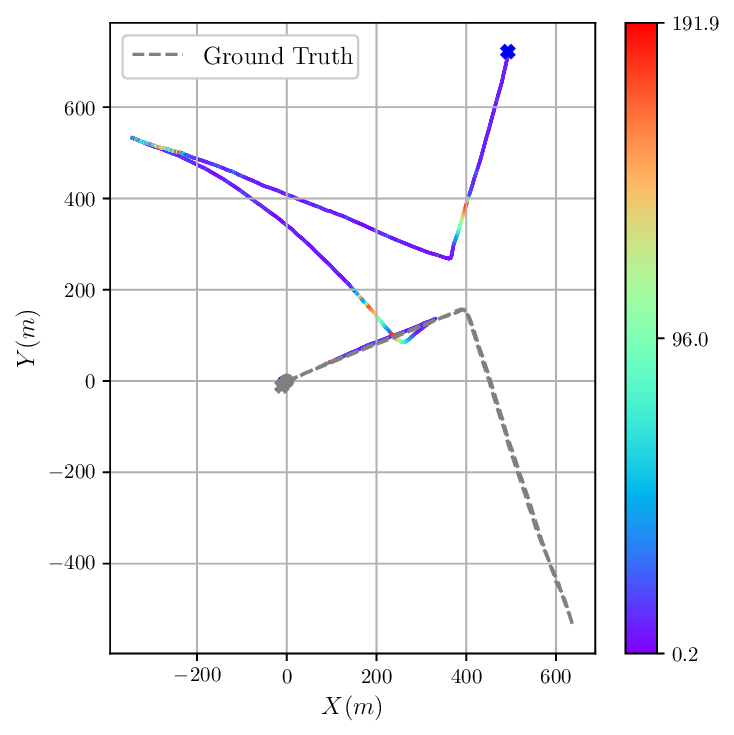} &
        \includegraphics[width=0.6\textwidth,height=2in]{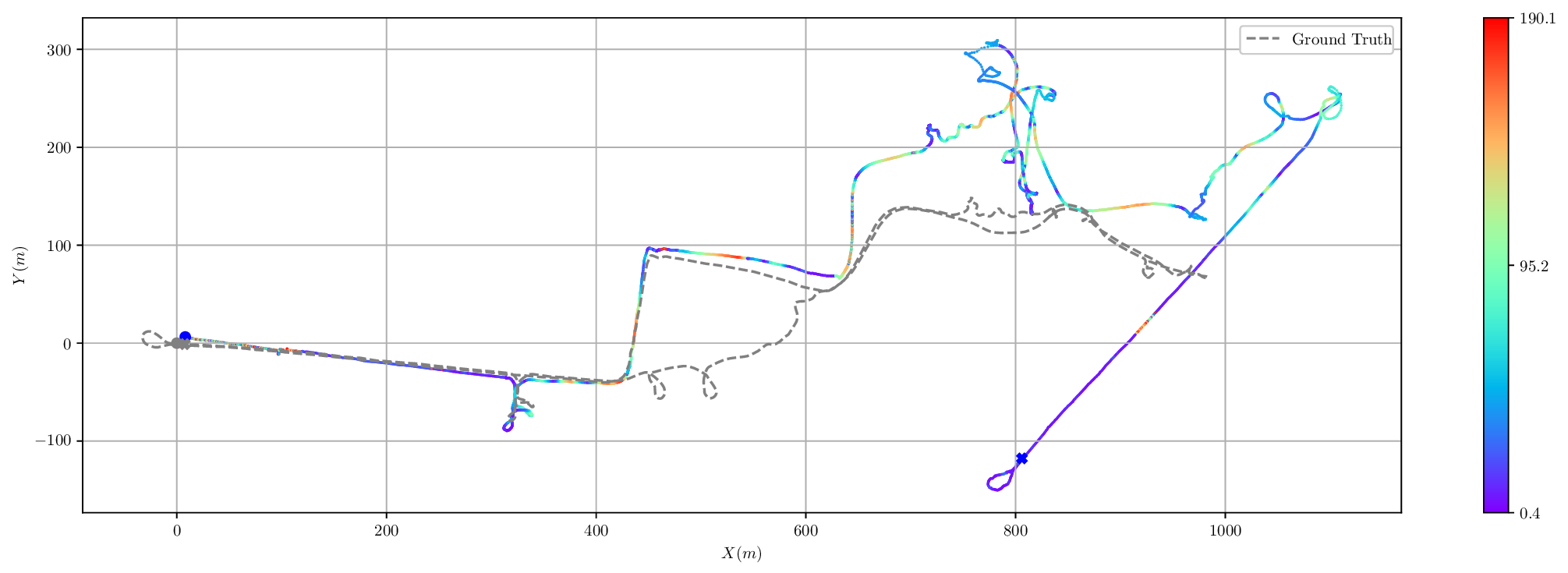} \\
        (a) Route 1 with trajectory from ORORA. 
        &  (b) Route 4 with trajectory from CFEAR. 
    \end{tabular}
    \caption{Trajectory plots with overlayed Relative Pose Error}
    \label{fig:rpe_overlayed}
\end{figure*}
\subsection{Comparative Insights}

From the combined results, several trends emerge:
\begin{itemize}
    \item \textbf{Planar vs. non-planar motion:} Both CFEAR and ORORA perform strongly in $SE(2)$, but degrade when pitch/roll are introduced.
    \item \textbf{Failure cases:} Ravine traversals expose the vulnerability of radar-only pipelines, with ground returns dominating scans and producing catastrophic drift.
    \item \textbf{Proposed Solutions:} Radar-KISSICP highlights the role of motion compensation in reducing planar bias, while Radar-IMU demonstrates that even simple IMU integration significantly stabilizes trajectory estimation in $SE(3)$.
\end{itemize}

Overall, these results emphasize that state-of-the-art radar odometry engines, though effective in structured environments, do not generalize reliably to unstructured terrain. Off-road performance requires either inertial integration or more robust approaches to handle 3D geometry and radar artifacts.

\section{Discussion}
\subsection{Urban Success Does Not Transfer to Off-road}
\begin{figure}[b]
    \centering
    \fbox{\includegraphics[width=\columnwidth]{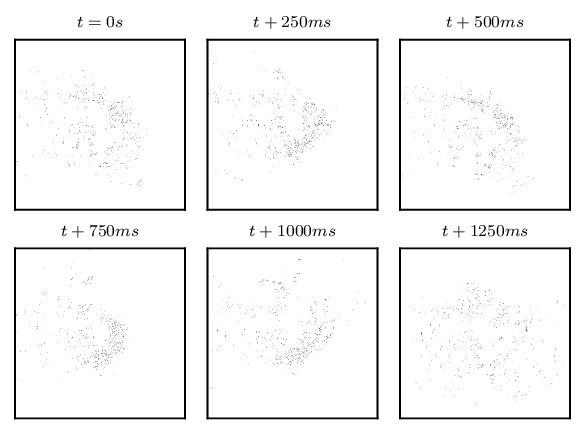}}
    \caption{Heavy pitching from portion of Route 0 in the GO dataset}
    \label{fig:pitching}
\end{figure}

Our experiments confirm that current radar odometry engines achieve strong performance on automotive datasets, where motion is largely constrained to $SE(2)$. In these settings, both CFEAR and ORORA estimate relative transforms reliably and produce trajectories with low drift. However, when applied to off-road sequences from the GO dataset, performance drops sharply. To better understand why current RADAR odometry methods show effectiveness while estimating relative transforms, but yield unfavorable trajectory estimates and therefore, higher absolute trajectory errors, we plot the relative pose error onto the trajectory in Figure \ref{fig:rpe_overlayed}. Here we can observe that except for two locations, the relative pose is accurately estimated. However, at these two locations, the error is significant enough to disturb the rest of the trajectory estimate. The RADAR measurements at this location are shown in Figure \ref{fig:pitching}, with Cen2018 as the feature detector. CFEAR shows robustness to a limit in this scenario, likely due to using scan-to-keyframe matching than sequential matching. In a scenario with more foliage and more deviations in to off-road areas like Route 4, these occurrences can be more catastrophic (Figure \ref{fig:rpe_overlayed} (b)).The difference stems from frequent pitch and roll excitations caused by rough terrain. In $SE(3)$ conditions, ground returns become a persistent part of the radar measurement rather than removable outliers, breaking the assumptions underlying urban automotive pipelines.

\begin{figure}
    \centering
    \includegraphics[width=\columnwidth]{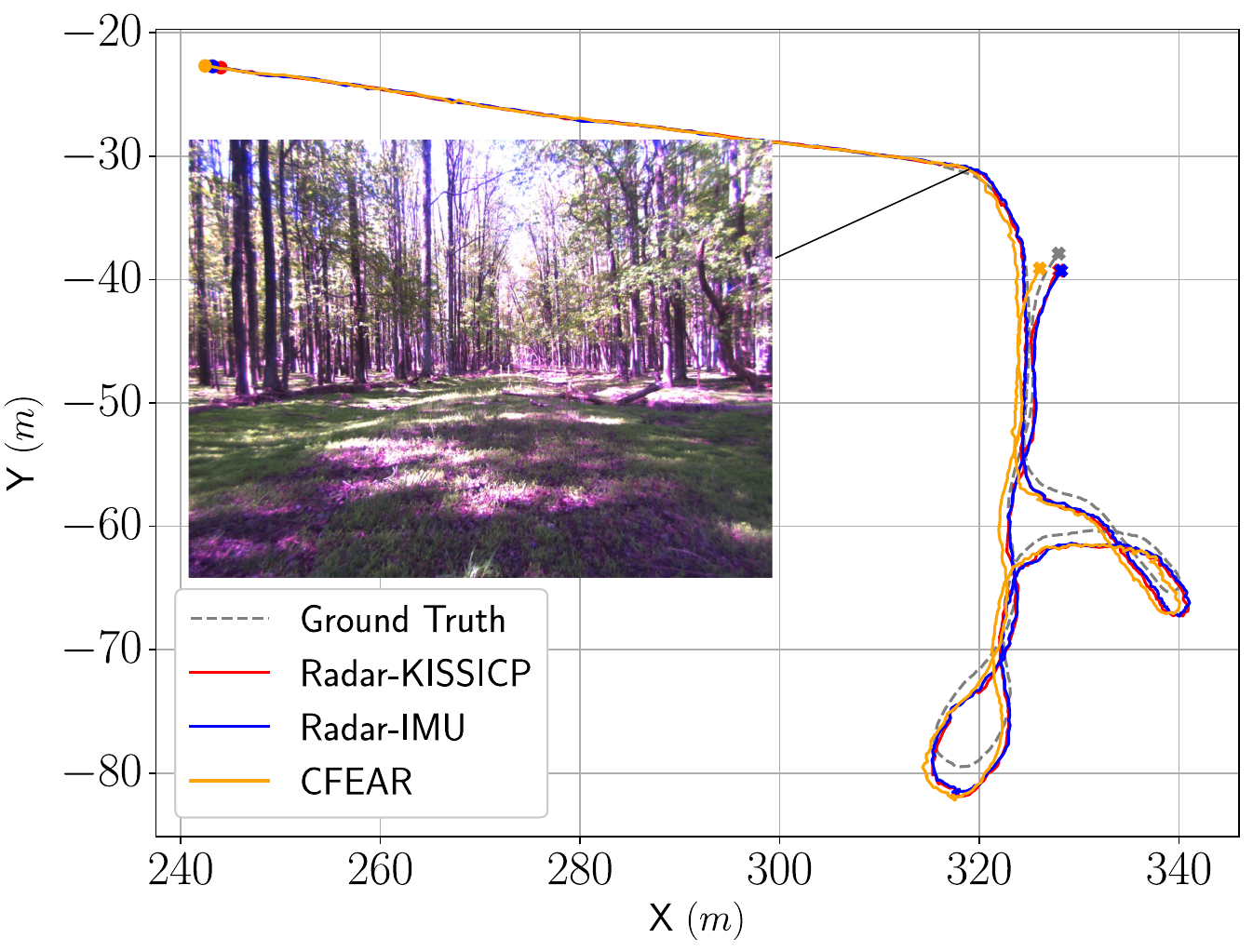}
    \caption[Isolated offroad sequence from Route 4]{%
    Trajectory plots from Radar-KISSICP, CFEAR, Radar-IMU with groundtruth, inset
    RGB image of the area traversed in this isolated sequence}
    \label{fig:isolated_r1r2} 
\end{figure}

The above effect is not isolated to offroad trajectories, and can be found in automotive sequences as well. From figures \ref{fig:planarity} we can see the differences in the planarness and hence the aggressiveness of motion. A common contribution comes from pitching of the vehicle which is found in both the automotive and offroad sequences. In the offroad sequences, this can occur due to terrain, and with the automotive datasets this can be seen when the vehicle stops or starts due to heave. The difference here lies in the rate and magnitude of this pitching. In the automotive sequences, both current engines incorporate some method to discard the effects of these ground returns. ORORA uses a graph-based pruning approach to discard false matches arising from these ground returns, and CFEAR uses a point-to-line error metric with keyframing to either discard associations and use better matching candidates. Additionally, we find that these methods are effective in offroad scenarios as well, while this pitching motion is within a certain limit. To support this, we show an isolated section from Route 4, with trajectories for CFEAR, Radar-KISSICP and Radar-IMU in Figure \ref{fig:isolated_r1r2}.

\subsection{Failure Modes}
The key consequence of $SE(3)$ motion is that ground returns can no longer be safely discarded as outliers. For CFEAR, which depends on consistent feature matching, failure begins when stable features are absent and ICP associations become ill-conditioned. This initiates a cycle of false matches, poor pose updates, and map corruption. ORORA avoids some of these issues by pruning outliers more aggressively, but in practice this can also remove useful matches, limiting robustness.

\subsection{Mitigation}
Our intention with Radar-KISSICP and Radar-IMU is to explore possible methods by which we can overcome these issues. Radar-KISSICP and Radar-IMU provide additional insight into these failure modes. Radar-KISSICP applies $SO(3)$ motion compensation by synchronizing radar sweeps with IMU orientations. This generates sparse but 3D-aware pointclouds that reduce the planar bias of standard pipelines. However, because these features depend on the radar pose, they are not spatially consistent, which leads to false feature accumulation in the map. 

\begin{figure}[!h]
    \centering
    \includegraphics[width=1\linewidth]{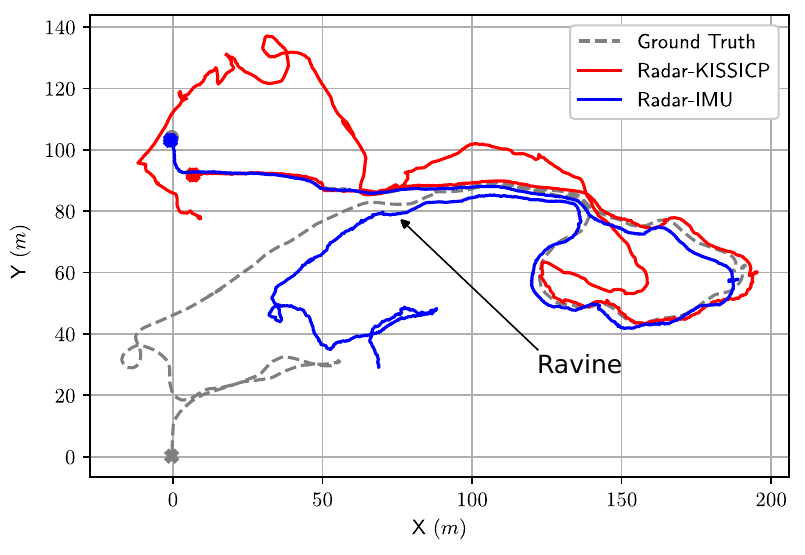}
    \caption{Isolated trajectory from Route 3 of the GO dataset. The black arrow shows the location of the ravine.}
    \label{fig:ravine_recovery}
\end{figure}

\begin{figure*}
    \centering
    \setlength{\tabcolsep}{0.2pt}
    \begin{tabular}{P{\columnwidth} P{\columnwidth}}
        \includegraphics[width=0.9\columnwidth,height=2in]{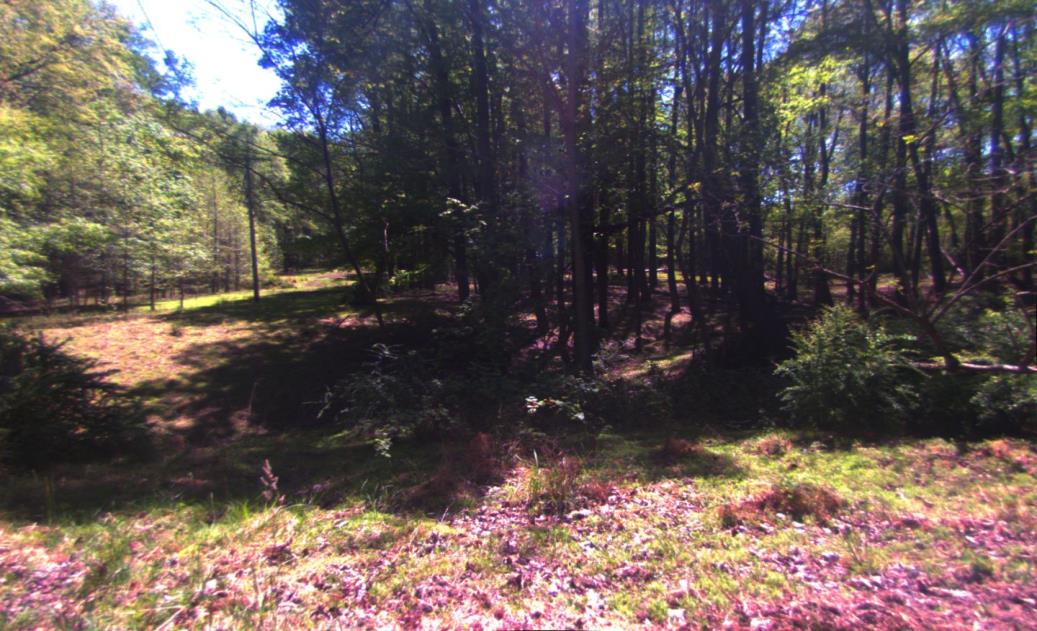} &
        \includegraphics[width=0.9\columnwidth,height=2in]{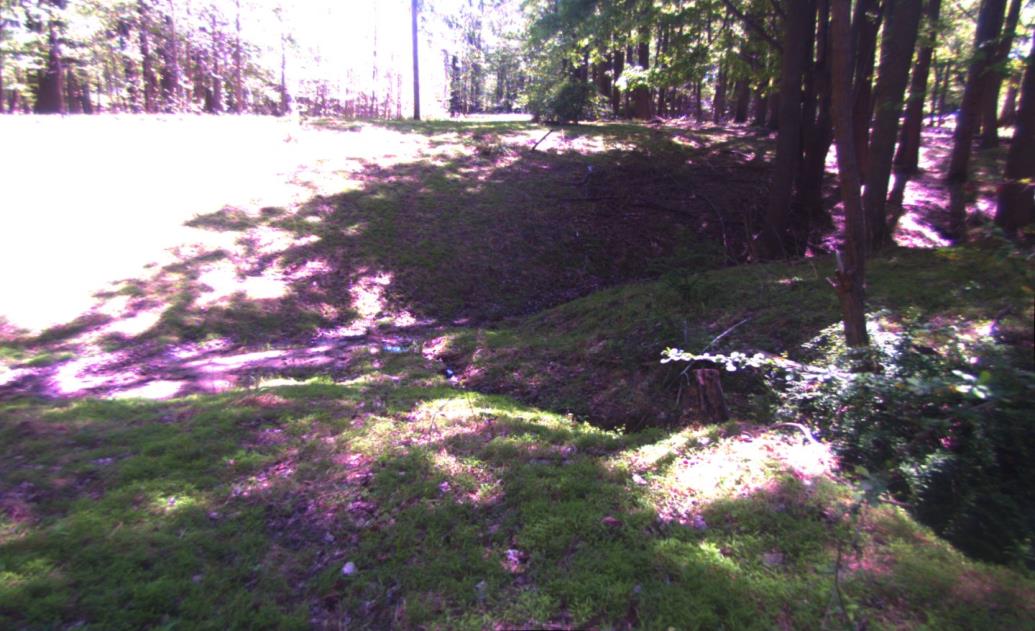} \\
        \includegraphics[width=0.9\columnwidth,height=2in]{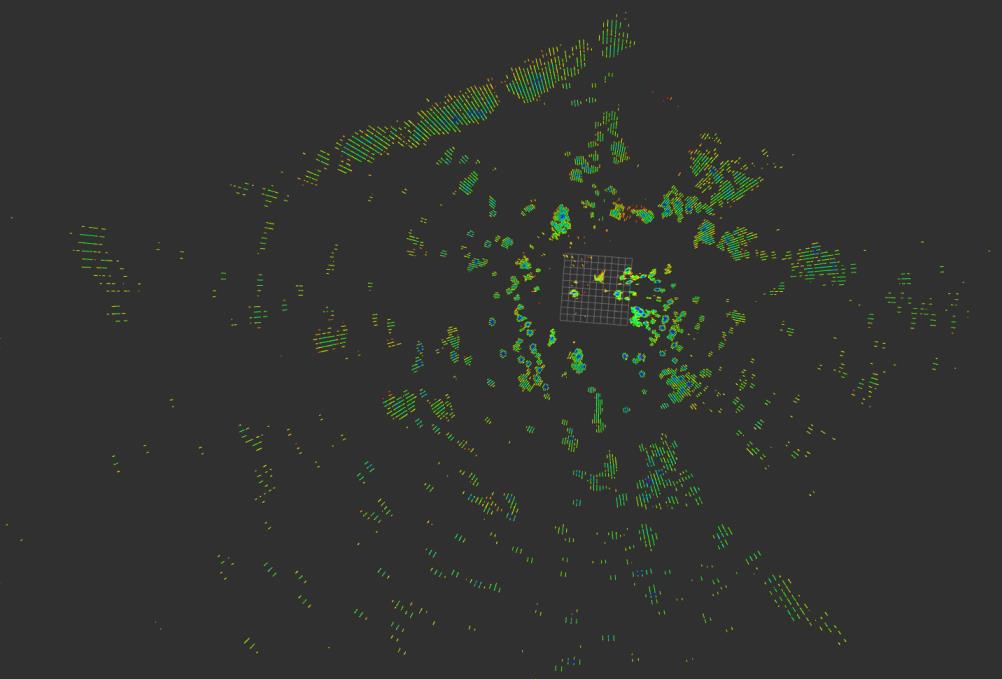} &
        \includegraphics[width=0.9\columnwidth,height=2in]{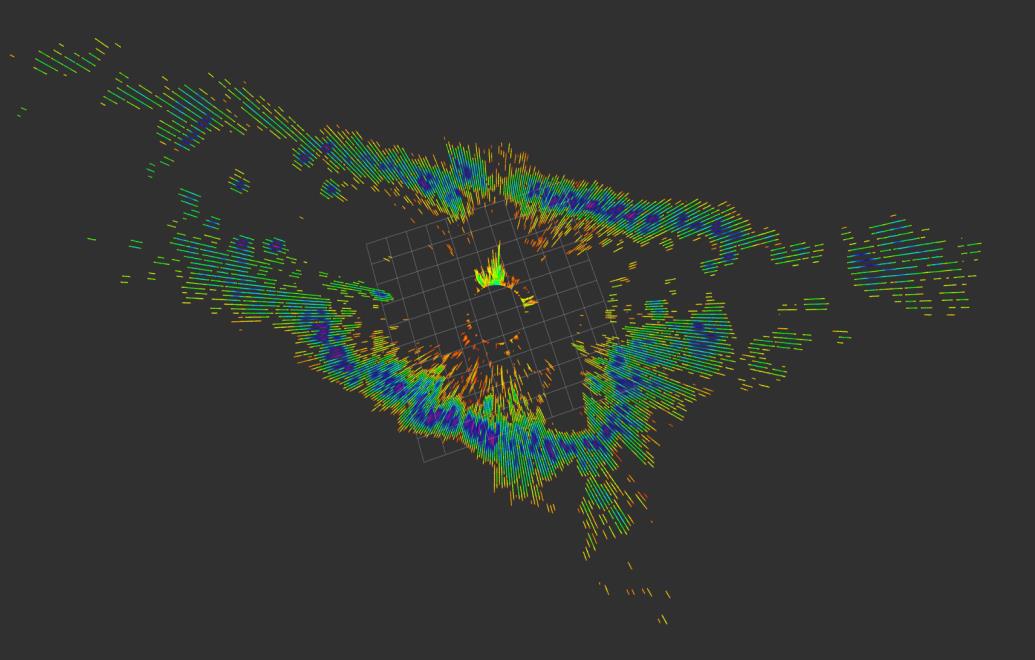} 
    \end{tabular}
    \caption{Camera images and radar pointclouds from Route 3.
    Left: The Warthog is on level ground at the edge of the ravine and the radar pointcloud is well populated and feature rich with surrounding trees. Right: As the warthog goes further into the ravine, these features are no longer visible until the warthog reaches the deepest point. Here, the radar returns are from the walls of the ravine, and none of the previous features are visible.}
    \label{fig:ravine}
\end{figure*}

This effect is visible in Route~3 during a ravine traversal. Before entering the ravine, many features are visible which are both spatially well dispersed and distinct by intensity. However, once the warthog enters the ravine, none of these features are visible, and the intensity returns from the radar are instead populated with reflections from the ravine walls (Fig.~\ref{fig:ravine}).  Without compensation, these returns project onto the same plane as prior scans and appear valid to ICP, compounding the drift. These would also lie well within the threshold for KNN matching, given the initial guess coming from a constant velocity model.

Radar-IMU mitigates this by introducing a full $SE(3)$ prior via IMU preintegration. Even without bias correction, the inertial estimate captures vertical displacement that radar alone cannot observe. This additional information helps reject false correspondences and improves recovery in difficult sections. In the same ravine sequence, Radar-IMU is able to recover where Radar-KISSICP diverges (Fig.~\ref{fig:ravine_recovery}), underscoring the importance of inertial cues in off-road radar odometry.

\subsection{Lessons Learned}
From these experiments, several lessons emerge. First, radar odometry methods that perform well in planar domains are brittle under $SE(3)$ terrain, where ground returns form an inseparable part of the signal. Second, isolated local errors can propagate to global drift, meaning robustness to rare but severe conditions is as important as average-case accuracy. Finally, our  highlight two promising directions: motion compensation to reduce planar assumptions, and inertial integration to stabilize trajectory estimation. Together, these findings point to the need for richer sensing (e.g., 3D radar) and tighter multi-sensor fusion to achieve reliable off-road localization.

\section{CONCLUSIONS}

In this paper, we presented a systematic evaluation of radar odometry in off-road environments. While state-of-the-art methods such as CFEAR and ORORA achieve near LiDAR-level accuracy on urban, largely planar datasets, our experiments demonstrate that their assumptions break down when extended to $SE(3)$ terrain. Through experiments on the Great Outdoors (GO) dataset, we showed that pitch/roll excitation, terrain-induced ground returns, and sparse features lead to significant trajectory drift, even when relative motion estimates remain locally accurate.

To better understand these limitations, we introduced two baselines. Radar-KISSICP incorporates an $SO(3)$ motion compensation step, revealing how planar assumptions contribute to false correspondences. Radar-IMU extends this by adding IMU preintegration, which provides vertical displacement awareness and recovers trajectories in cases where radar-only pipelines fail. While neither baseline is intended as a new algorithm, both highlight critical factors for robust radar odometry in unstructured terrain.

From this evaluation, we draw several lessons. First, ground returns in off-road scenarios must be treated as integral parts of the measurement rather than removable outliers. Second, robustness to isolated but severe failures is as important as average-case accuracy, since small local errors can propagate to catastrophic global drift. Finally, inertial cues and richer sensing modalities are essential to bridge the gap between radar’s robustness to adverse weather and the geometric complexity of natural environments.

We believe this work provides both a benchmark and a diagnostic perspective for future research in off-road radar localization. Promising directions include the use of 3D radar, learned denoising and feature selection, and tighter multi-sensor fusion with IMU, LiDAR and visual/LWIR cameras. By identifying and analyzing the key failure modes of current methods, our study lays the groundwork for more resilient radar-based localization in unstructured environments.





\section*{ACKNOWLEDGMENT}
Research reported in this paper was sponsored in part by the DEVCOM Army Research Laboratory under Cooperative Agreements W911NF-24-2-0023 (SARA CRA) and W911NF-23-2-0211. The views and conclusions contained in this document are those of the authors and should not be interpreted as representing the official policies, either expressed or implied, of the DEVCOM Army Research Laboratory or the U.S. Government. The U.S. Government is authorized to reproduce and distribute reprints for Government purposes, notwithstanding any copyright notation herein.

\bibliographystyle{IEEEtran}
\bibliography{IEEEabrv, offroad_radar_odometry}

\end{document}